\documentclass[conference]{IEEEtran}
\IEEEoverridecommandlockouts
\usepackage{cite}
\usepackage{amsmath,amssymb,amsfonts}
\usepackage{blindtext}
\usepackage{algorithmic}
\usepackage{graphicx}
\usepackage{textcomp}
\usepackage{xcolor}
\usepackage{multirow}
\usepackage{subfiles}
\usepackage{bbm}
\usepackage[super]{nth}
\def\BibTeX{{\rm B\kern-.05em{\sc i\kern-.025em b}\kern-.08em
    T\kern-.1667em\lower.7ex\hbox{E}\kern-.125emX}}
\begin{document}

\title{CropCat: Data Augmentation for Smoothing the Feature Distribution of EEG Signals
\footnote{{\thanks{This research was supported by the Challengeable Future Defense Technology Research and Development Program (912911601) of Agency for Defense Development in 2020.}
}}
}

\author{\IEEEauthorblockN{Sung-Jin Kim}
\IEEEauthorblockA{\textit{Dept. Artificial Intelligence} \\
\textit{Korea University} \\
Seoul, Republic of Korea \\
s\_j\_kim@korea.ac.kr} \\

\and

\IEEEauthorblockN{Dae-Hyeok Lee}
\IEEEauthorblockA{\textit{Dept. Brain and Cognitive Engineering} \\
\textit{Korea University} \\
Seoul, Republic of Korea \\
lee\_dh@korea.ac.kr} \\

\and

\IEEEauthorblockN{Yeon-Woo Choi}
\IEEEauthorblockA{\textit{Dept. Artificial Intelligence} \\
\textit{Korea University} \\
Seoul, Republic of Korea \\
yw\_choi@korea.ac.kr}


}

\maketitle

\begin{abstract}
Brain-computer interface (BCI) is a communication system between humans and computers reflecting human intention without using a physical control device. Since deep learning is robust in extracting features from data, research on decoding electroencephalograms by applying deep learning has progressed in the BCI domain. However, the application of deep learning in the BCI domain has issues with a lack of data and overconfidence. To solve these issues, we proposed a novel data augmentation method, CropCat. CropCat consists of two versions, CropCat-spatial and CropCat-temporal. We designed our method by concatenating the cropped data after cropping the data, which have different labels in spatial and temporal axes. In addition, we adjusted the label based on the ratio of cropped length. As a result, the generated data from our proposed method assisted in revising the ambiguous decision boundary into apparent caused by a lack of data. Due to the effectiveness of the proposed method, the performance of the four EEG signal decoding models is improved in two motor imagery public datasets compared to when the proposed method is not applied. Hence, we demonstrate that generated data by CropCat smooths the feature distribution of EEG signals when training the model.

\end{abstract}

\begin{small}
\end{small}

\section{INTRODUCTION}

Brain-computer interface (BCI) is a system that makes humans could control the computer based on human intentions without physical interactions by linking the brain and computer \cite{suk2014predicting, zhang2017hybrid, won2017motion, zhang2021adaptive, lee2019connectivity, thung2018conversion, jeong2020decoding}. In particular, an electroencephalogram (EEG)-based BCI system is extensively studied due to its high portability and practicality. The ability to control the computer without physical actions is an effective system for communicating with patients with difficulty moving their bodies, like those with a stroke. In addition to the rehabilitation perspective, the capacity to control the computer reflecting human intentions could enhance safety and productivity in the industry aspect. Therefore, the BCI system has been studied in various applications, such as a robotic arm, speller, drone, and wheelchair \cite{lee2018high, yu2018asynchronous, kim2019subject, lee2020continuous, jeong2020brain, cho2021neurograsp, lee2021subject}.

The superior decoding performance of the model is necessary to construct the EEG-based BCI systems. As the neural network-based deep learning model showed remarkable performance by regressing the general distribution derived from many data based on gradient descent, the deep learning model has also been studied in the BCI domain to decode EEG signals \cite{ang2008filter, schirrmeister2017deep, kwon2019subject, lawhern2018eegnet, kim2022rethinking}. However, some issues exist in applying deep learning in BCI. The first is the lack of data issue. The deep learning model needs abundant data to find the optimal point. On the contrary, since EEG is a biological signal, acquiring data is difficult. The great difficulty of tasks, like imagination tasks, also makes data acquisition harder. The second is the overconfidence issue. Deep learning traditionally has the overconfidence issue. In the BCI domain, since the number of data is small, the deep learning model rapidly overfits the training data and makes more overconfidence in its predictions.

These two issues are all derived from the lack of data. Therefore, much research has been studied to solve this issue using data augmentation. Cheng \textit{et al.} \cite{cheng2020subject} conducted ten data augmentation methods at EEG signals to train the self-supervised-based model, which needs abundant data to search for the optimal pre-trained model. Although this approach alleviated the lack of data issue, the overconfidence issue of the deep learning model still exists. Zhang \textit{et al.} \cite{zhang2022eeg} proposed the generative adversarial network-based data augmentation method to improve the classification performance in the EEG signals-based emotion recognition domain. They enhanced the performance by generating unseen data using Wasserstein generative adversarial network (GAN). However, since the quantity of actual data is small, the issue that the classifier might learn artificially generated features from GAN, which are not real, in a biased way exists.

To solve the lack of data and overconfidence issues, we proposed the novel data augmentation method, CropCat, which augments data by cropping and concatenating the different class data of each subject. We successfully generated the data originating from real data and improved the classification performance training with CropCat. In addition, we refined the overconfidence issue by smoothing the label based on the class ratio in each data.

\section{MATERIALS AND METHODS}
\subsection{Datasets}

We used two open datasets to evaluate our method, BCI Competition IV 2a (dataset 2a) \cite{brunner2008bci} and BCI Competition IV 2b (dataset 2b) \cite{leeb2008bci}. These datasets are the most used public datasets to evaluate the performance of EEG signal decoding methods \cite{al2021deep}. We conducted the low-pass filtering at 38 Hz to only leave the frequency relevant to MI (mu and beta rhythms). \cite{nicolas2012brain, hobson2017interpretation} In addition, exponential moving standardization was performed to remove the peaks irrelevant to MI tasks. \cite{kim2022rethinking}



\subsection{CropCat}

The previous data augmentation studies \cite{banville2021uncovering, jiang2021self, zhang2022ganser} were focused on deletion, the addition of noise, and generation. These approaches showed significant performance in visually confirmable domains, like computer vision. In the case of EEG signals that could not visually confirm the change, since it is difficult to verify even if the signals are damaged through the data augmentations, the issue exists to apply these methods. Hence, we focused on generating the data based on real signals.

The EEG signals are the biosignals acquired from brain activity. The EEG signals are measured by calculating the voltage difference between the active and ground electrodes. Through this process, we could get spatial and temporal information on brain activity. Since spatial and temporal features are the main characteristics of EEG signals measuring from this process, we designed our proposed data augmentation method to enrich the spatial and temporal information.

The dataset consisted of pair containing one EEG data and label. We selected one pair and set this pair to the base (base pair) ($(X_b, y_b)\in{B}, X_b \in \mathbbm{R}^{C \times T}, y_b \in \mathbbm{R}$) for applying the data augmentation. After choosing the base data pair, we sample another data pair (material pair) ($(X_m, y_m)\in{B}, X_m \in \mathbbm{R}^{C \times T}, y_m \in \mathbbm{R}, y_m \neq y_b$) in the same subject whose label differs from the base data pair. For ease of calculation, since the batch size is sufficiently large, we sampled the material pair from the same mini-batch of the base pair. We denote ${B}$ as the mini-batch, ${C}$ as the the number of channels, and ${T}$ as the number of time points. In addition, $X_b$ and $X_m$ are the EEG signals, and $y_b$ and $y_m$ are the labels of the corresponding data. In addition, we could express $X_b$ and $X_m$ as follows:

\vspace{-0.3cm}
\begin{equation}
\begin{split}
    &{X_b} = [ c_{b1} ; c_{b2} ; \cdots ; c_{bC} ], \quad c_{bi} \in \mathbbm{R}^{1 \times T}, \; 1 \leq i \leq C \\
    &\quad\; = [ t_{b1}, t_{b2}, \cdots, t_{bT} ], \quad t_{bj} \in \mathbbm{R}^{C \times 1}, \; 1 \leq j \leq T \\
    &{X_m} = [ c_{m1} ; c_{m2} ; \cdots ; c_{mC} ], \quad c_{mi} \in \mathbbm{R}^{1 \times T}, \; 1 \leq i \leq C \\
    &\quad\;\; = [ t_{m1}, t_{m2}, \cdots, t_{mT} ], \quad t_{mj} \in \mathbbm{R}^{C \times 1}, \; 1 \leq j \leq T
\end{split}
\end{equation}

We designed the proposed method by dividing the spatial (CropCat-spatial) and temporal (CropCat-temporal) ways. CropCat-spatial could mix the spatial information of two different labeled data. It might improve the decoding performance of ambiguous data, like simultaneously imagining the left and right hand at the same time point. CropCat-temporal could fuse the temporal information of two data. Using the augmented data through CropCat-temporal, the model could learn the ambiguous data, like imagining the left hand's movement for two sec. and then imagining the right hand's movement for one sec. when the left-hand task is assigned.

For applying CropCat-spatial and CropCat-temporal, we set the center point $c$, the anchor for mixing two data, and the $r$, which decides the ratio of using material pair. $c_s$ (the center point of CropCat-spatial) and $c_t$ (the center point of CropCat-temporal) are sampled from the uniform distribution.

\vspace{-0.3cm}
\begin{equation} 
\begin{split}
    &{c_s} \sim \text{Unif} (0, {C}) \\
    &{c_t} \sim \text{Unif} (0, {T})
\end{split}
\end{equation}

In addition, the mixing ratio $r$ is randomly sampled from the uniform distribution. We restricted the interval of uniform distribution with the max value in half to keep the effect of the base pair more robust. 

\vspace{-0.3cm}
\begin{equation} 
\begin{split}
    &{r_s}, {r_t} \sim \text{Unif} (0, \lambda), \quad \lambda \in [0, 0.5]
\end{split}
\end{equation}
where $r_s$ and $r_t$ indicate the ratio of CropCat-spatial and CropCat-temporal, respectively. In CropCat-spatial, we set the $\lambda$ as 0.333 in both dataset. In CropCat-temporal, we set the $\lambda$ as 0.125 in dataset 2a and 0.1 in dataset 2b.

We mixed the base and material data by conducting the cropping and concatenating processes based on set hyperparameters. In the case of CropCat-spatial, two data mixed the spatial information as follows the formula:

\vspace{-0.3cm}
\begin{equation} 
\begin{split}
    &\tilde{X}_{spatial} = {X}_{b}\mathbbm{1}_{t \ \in \ [1, \ c_s \ - \ rC/2) \ \cup \ (c_s \ + \ rC/2, \ C]} \\ 
    &\quad\quad\quad\quad\quad\quad\quad\quad\quad + {X}_{m}\mathbbm{1}_{t \ \in \ [c_s \ - \ rC/2, \ c_s \ + \ rC/2]} \\
    &\quad\quad\quad\;\,=[ c_{b1} ; c_{b2} ; \cdots ; c_{b(c_s-rC/2-1)} ; c_{m(c_s-rC/2)}; \\ 
    &\quad\quad\quad\quad\quad \cdots ; c_{m(c_s+rC/2)} ; c_{b(c_s-rC/2+1)} ; \cdots ; c_{bC} ]
\end{split}
\end{equation}

In addition, in the case of CropCat-temporal, both data fused the temporal information as follows the formula:

\vspace{-0.3cm}
\begin{equation} 
\begin{split}
    &\tilde{X}_{temporal} = {X}_{b}\mathbbm{1}_{t \ \in \ [1, \ c_t \ - \ rT/2) \ \cup \ (c_t \ + \ rT/2, \ T]} \\ 
    &\quad\quad\quad\quad\quad\quad\quad\quad\quad + {X}_{m}\mathbbm{1}_{t \ \in \ [c_t \ - \ rT/2, \ c_t \ + \ rT/2]} \\
    &\quad\quad\quad\quad\,= [ t_{b1}, t_{b2}, \cdots, t_{b(c_t-rT/2-1)}, t_{m(c_t-rT/2)}, \\
    &\quad\quad\quad\quad\quad\quad \cdots, t_{m(c_t+rT/2)}, t_{b(c_t+rT/2+1)}, \cdots, t_{bT} ]
\end{split}
\end{equation}
where $\tilde{X}_{spatial}, \tilde{X}_{temporal} \in\mathbbm{R}^{{C}\times{T}}$ are the mixed data in spatial and temporal ways, respectively.

Since the base and material data are mixed, we generated a fused label ($\tilde{y}$) based on the ratio applied to the data. The detailed formula is as follows:

\vspace{-0.3cm}
\begin{equation} 
\begin{split}
    \tilde{y} = (1 \ - \ {r}){y_b} \ + \ {r}{y_m}, \quad \tilde{y}\in\mathbbm{R}
\end{split}
\end{equation}

Based on CropCat-spatial and CropCat-temporal, we generated novel data containing various spatial and temporal information using real data. We trained the decoding models by including these augmented data pairs $(\tilde{X}, \tilde{y})$ in the training data.


\begin{table*}[t!]
\centering
\caption{Comparison of performances for decoding EEG signals in four different EEG decoding models using datasets 2a and 2b applying the conventional data augmentation methods and the proposed method.}
\renewcommand{\arraystretch}{1.2}
\setlength{\arrayrulewidth}{0.13mm}
\tiny
\resizebox{\textwidth}{!}{
\begin{tabular}{l|cc|l|cc}
\hline
\multicolumn{1}{c|}{\multirow{2}{*}{Method}} & Dataset 2a           & Dataset 2b           & \multicolumn{1}{c|}{\multirow{2}{*}{Method}} & Dataset 2a           & Dataset 2b           \\ \cline{2-3} \cline{5-6} 
\multicolumn{1}{c|}{}                        & Accuracy$\pm$std.      & Accuracy$\pm$std.      & \multicolumn{1}{c|}{}                        & Accuracy$\pm$std.      & Accuracy$\pm$std.      \\ \hline
ShallowConvNet\cite{schirrmeister2017deep}                               & 0.781$\pm$0.005          & 0.841$\pm$0.008          & DeepConvNet\cite{schirrmeister2017deep}                                  & 0.714$\pm$0.010          & 0.833$\pm$0.002          \\
+ Time masking                               & 0.628$\pm$0.004          & 0.818$\pm$0.007          & + Time masking                               & 0.608$\pm$0.003          & 0.786$\pm$0.005          \\
+ Gaussian noise                             & \textbf{0.785$\pm$0.006} & 0.835$\pm$0.005          & + Gaussian noise                             & \textbf{0.715$\pm$0.004} & \textbf{0.843$\pm$0.005} \\
+ CutOut\cite{devries2017improved}                                     & 0.745$\pm$0.006          & \textbf{0.860$\pm$0.001} & + CutOut\cite{devries2017improved}                                     & 0.645$\pm$0.004          & 0.831$\pm$0.003          \\
+ CropCat-spatial                            & 0.694$\pm$0.005          & 0.827$\pm$0.002          & + CropCat-spatial                            & 0.483$\pm$0.002          & 0.681$\pm$0.005          \\
+ CropCat-temporal                           & 0.783$\pm$0.005          & \textbf{0.860$\pm$0.004} & + CropCat-temporal                           & \textbf{0.715$\pm$0.005} & \textbf{0.843$\pm$0.005} \\ \hline
\multicolumn{1}{c|}{\multirow{2}{*}{Method}} & Dataset 2a           & Dataset 2b           & \multicolumn{1}{c|}{\multirow{2}{*}{Method}} & Dataset 2a           & Dataset 2b           \\ \cline{2-3} \cline{5-6} 
\multicolumn{1}{c|}{}                        & Accuracy$\pm$std.      & Accuracy$\pm$std.      & \multicolumn{1}{c|}{}                        & Accuracy$\pm$std.      & Accuracy$\pm$std.      \\ \hline
EEGNet\cite{lawhern2018eegnet}                                       & 0.730$\pm$0.002          & 0.825$\pm$0.009          & M-ShallowConvNet\cite{kim2022rethinking}                             & 0.809$\pm$0.004          & 0.861$\pm$0.002          \\
+ Time masking                               & 0.639$\pm$0.026          & 0.796$\pm$0.017          & + Time masking                               & 0.686$\pm$0.003          & 0.841$\pm$0.007          \\
+ Gaussian noise                             & \textbf{0.734$\pm$0.011} & 0.830$\pm$0.010          & + Gaussian noise                             & 0.804$\pm$0.010          & 0.860$\pm$0.002          \\
+ CutOut\cite{devries2017improved}                                     & 0.708$\pm$0.017          & 0.835$\pm$0.014          & + CutOut\cite{devries2017improved}                                     & 0.776$\pm$0.003          & 0.861$\pm$0.002          \\
+ CropCat-spatial                            & 0.622$\pm$0.001          & 0.695$\pm$0.011          & + CropCat-spatial                            & 0.689$\pm$0.006          & 0.836$\pm$0.002          \\
+ CropCat-temporal                           & 0.730$\pm$0.007          & \textbf{0.835$\pm$0.019} & + CropCat-temporal                           & \textbf{0.811$\pm$0.004} & \textbf{0.863$\pm$0.003} \\ \hline
\end{tabular}}
\end{table*}

\subsection{Evaluation Settings}

We used three conventional data augmentation methods to compare the performances of methods, time masking, Gaussian noise, and CutOut \cite{devries2017improved}. Time masking is one of the traditional data augmentation methods which masks the specific interval of data. We set the masking ratio as 0.1 and 0.05 in dataset 2a and 2b, respectively. Adding Gaussian noise is a common data augmentation that could apply in various domains. We set the mean and standard deviation as zero and one, respectively. In addition, we scaled the noise by multiplying 0.05 before adding the noise. CutOut is the data augmentation method firstly proposed in the computer vision domain. This method generates the data by removing some random regions in the data. We set the channel lengths, time lengths, and the number of regions to remove as 0.25, 0.5, and 3, respectively.

We selected three models which are the most commonly used, ShallowConvNe, DeepConvNet, and EEGNet. In addition, we selected M-ShallowConvNet, which refined the issues that arose in ShallowConvNet. We trained the models for 1,000 epochs as a batch size of 64. Adam optimizer was used as a learning rate of 2e-3, and the cosine learning rate scheduler was applied to improve the training stability \cite{loshchilov2016sgdr}. The loss function was set to cross-entropy loss. We saved the checkpoint, which shows the lowest loss value, and used this checkpoint as the final model. Since the data size is small, we applied five-fold cross-validation to train the model. After training the five models by conducting five-fold cross-validation, we decided on results by voting. For the evaluation metrics, we used accuracy and standard deviation (std.).

\section{RESULTS AND DISCUSSION}

Table 1 presents the decoding performances of four baseline models trained using various conventional data augmentation methods (time masking, Gaussian noise, and CutOut) and the proposed methods (CropCat-spatial and CropCat-temporal) in datasets 2a and 2b. As shown in the Table 1, when conducting the time masking, all decoding performances are mostly dropped in both datasets compared to when data augmentations are not applied. Since the essential features of MI-based EEG signals are leaned in the specific region, augmented data quality is decreased when the critical parts are masked. In the case of adding Gaussian noise, the performances of ShallowConvNet, DeepConvNet, and EEGNet in dataset 2a and DeepConvNet in dataset 2b achieved the best accuracy. However, the performances are dropped compared to the baseline in other cases. Based on these results, we thought that a fine hyperparameter optimization process was needed for applying Gaussian noise to EEG signals where visual features could not be identified. CutOut shows the different aspects depending on the dataset. Although the performances of all models in dataset 2a are dropped compared to the baseline, the performances of ShallowConvNet and EEGNet are improved. We thought that, unlike time masking, CutOut caused performance improvements in both models because it only obscured some information, not all information, at certain times. However, since the performance improvement occurred only in some cases, we thought optimizing the hyperparameters was difficult, similar to Gaussian noise. In addition, we verified that the performances in four models when conducting CropCat-spatial are significantly dropped. We thought that since the spatial information of EEG signals at a specific time is important, the spatially fused data by CropCat-spatial generates data that have distorted information. However, in the case of CropCat-temporal, decoding performances of all models are improved compared to the baseline with 0.783, 0.715, 0.730, and 0.811 in dataset 2a and 0.860, 0.843, 0.835, and 0.863 in dataset 2b, respectively. Since CropCat-temporal produced data that could occur by generating the data based on real EEG signals, we could solve the lack of data issue in the BCI domain.



\section{CONCLUSION AND FUTURE WORKS}

In this paper, we proposed the novel data augmentation method, CropCat, which could apply in the BCI domain to decode EEG signals. Some issues with applying deep learning exist in the BCI domain. The first is the lack of data issue. Since the EEG signals are biosignals, the data acquisition cost is high. Therefore, acquiring sufficient data to train the deep learning-based model is difficult. The second is the overconfidence issue of the deep learning model. The overconfidence issue is the traditional issue in deep learning-based methods. In addition, the lack of data in the BCI domain exacerbates the issue. To solve these issues, we designed the data augmentation method, which fuses the two real EEG signals having different labels. Since the labels were adjusted according to the percentage of the data combined, the sharp distribution of data's features converted smoother. As a result, we improved the decoding performances of ShallowConvNet, DeepConvNet, and M-ShallowConvNet in dataset 2a with 0.783, 0.715, and 0.811, respectively. In addition, the decoding performances of ShallowConvNet, DeepConvNet, EEGNet, and M-ShallowConvNet are improved in dataset 2b with 0.860, 0.843, 0.835, and 0.863, respectively. We demonstrated with Grad-CAM that the model focuses on important features in the augmented data using CropCat, and we verified that the inference results of the deep learning model are not overconfident in augmented data. In future works, we will study the self-supervised learning algorithm for that the data augmentation method is critical. Hence, we will construct a pre-trained model that could be commonly used in various EEG signals.

\bibliographystyle{IEEEtran}
\bibliography{REFERENCE}

\end{document}